\documentclass[runningheads]{llncs}
\usepackage[T1]{fontenc}
\usepackage[english]{babel}
\usepackage{algorithm}
\usepackage{amsmath,amssymb}
\usepackage{algorithmic}
\usepackage{graphicx}
\usepackage{subcaption}
\usepackage{textcomp}
\usepackage{xcolor}
\usepackage{hyperref}
\usepackage{paralist}
\usepackage{booktabs} 
\usepackage{multirow}
\usepackage{bbding}
\usepackage[font=small,labelfont=bf]{caption}
\usepackage{orcidlink}
\usepackage[moderate]{savetrees}
\addto\extrasenglish{%
}

\begin{document}
\title{Neuro-Symbolic Learning for Predictive Process Monitoring via Two-Stage Logic Tensor Networks with Rule Pruning}
\titlerunning{Neuro-Symbolic Learning for Predictive Process Monitoring}
\author{Fabrizio De Santis \inst{1}\Envelope\orcidlink{0009-0002-5079-3048} \and
Gyunam Park\inst{2}\orcidlink{0000-0001-9394-6513} \and Francesco Zanichelli\inst{1}\orcidlink{0000-0002-5802-8343}}
\authorrunning{F. De Santis et al.}
\institute{University of Parma, Parma, Italy\\\email{\{fabrizio.desantis,francesco.zanichelli\}@unipr.it} \and
Eindhoven University of Technology, Eindhoven, Netherlands\\
\email{g.park@tue.nl}}
\maketitle              % typeset the header of the contribution
\begin{abstract}
Predictive modeling on sequential event data is critical for fraud detection and healthcare monitoring. Existing data-driven approaches learn correlations from historical data but fail to incorporate domain-specific sequential constraints and logical rules governing event relationships, limiting accuracy and regulatory compliance. For example, healthcare procedures must follow specific sequences, and financial transactions must adhere to compliance rules. We present a neuro-symbolic approach integrating domain knowledge as differentiable logical constraints using Logic Networks (LTNs). We formalize control-flow, temporal, and payload knowledge using Linear Temporal Logic and first-order logic. Our key contribution is a two-stage optimization strategy addressing LTNs' tendency to satisfy logical formulas at the expense of predictive accuracy. The approach uses weighted axiom loss during pretraining to prioritize data learning, followed by rule pruning that retains only consistent, contributive axioms based on satisfaction dynamics. Evaluation on four real-world event logs shows that domain knowledge injection significantly improves predictive performance, with the two-stage optimization proving essential knowledge (without it, knowledge can severely degrade performance). The approach excels particularly in compliance-constrained scenarios with limited compliant training examples, achieving superior performance compared to purely data-driven baselines while ensuring adherence to domain constraints.
\keywords{Predictive process monitoring, Neuro-symbolic AI, Logic Tensor Networks, Deep learning.}
\end{abstract}
\section{Introduction}
Modern organizations across finance, healthcare, and manufacturing routinely collect vast amounts of event data that are timestamped, attribute-rich records that capture the execution of business processes~\cite{DBLP:conf/kdd/HuangXH25}. 
Predictive process monitoring has emerged as a critical capability for these organizations, enabling them to anticipate process outcomes, detect anomalies, and optimize operational efficiency by analyzing historical event sequences~\cite{digh_2022,DBLP:conf/caise/TaxVRD17}. 

Recent advances in machine learning have improved predictive process monitoring capabilities. 
LSTM networks~\cite{DBLP:conf/caise/TaxVRD17,DBLP:journals/dss/EvermannRF17} and Transformer architectures~\cite{DBLP:journals/corr/abs-2104-00721} have shown superior performance in capturing complex sequential patterns and long-range dependencies in process data. 
Advanced approaches like bidirectional LSTM~\cite{DBLP:conf/bpm/0001DR19,DBLP:conf/icws/WangYLS19} and topological deep learning methods~\cite{DBLP:conf/kdd/HuangXH25} have further pushed the boundaries by incorporating richer structural representations and higher-order relationships between process events.

However, purely data-driven approaches face a fundamental limitation: they struggle to incorporate explicit domain constraints and business rules that are critical for real-world deployment~\cite{DBLP:conf/icml/XuZFLB18,DBLP:conf/icml/FischerBDGZV19}. 
In many domains, these rules represent mandatory compliance requirements or safety constraints that cannot be violated, even if historical data suggests otherwise. For example, in healthcare, ``surgery can only be scheduled if a patient was discharged more than seven days ago,'' or in banking, ``no transfer to a new beneficiary can occur before identity verification is complete.'' Such constraints may be rarely violated in training data yet remain absolutely critical during live operation. This disconnect between learned patterns and domain requirements highlights a key gap in current predictive process monitoring approaches.

\textit{Neuro-symbolic artificial intelligence} offers a promising solution by integrating neural networks' pattern recognition capabilities with symbolic reasoning's explicit knowledge representation~\cite{bhrato_2024,bagase_2022}. 
This paradigm enables models to leverage both statistical patterns from data and formal logical constraints, ensuring predictions that are both accurate and compliant with domain requirements. Recent frameworks like Logic Tensor Networks~\cite{bagase_2022}, semantic loss functions~\cite{DBLP:conf/icml/XuZFLB18}, and constraint-aware training methods~\cite{DBLP:conf/icml/FischerBDGZV19} have demonstrated the potential of neuro-symbolic approaches across various domains, yet their systematic application to predictive process monitoring remains largely unexplored.

This paper presents a comprehensive neuro-symbolic framework for predictive process monitoring that systematically integrates domain knowledge with neural learning. Our approach addresses three key challenges: (1) how to formalize diverse types of process constraints using logical representations, (2) how to handle conflicts between domain rules and historical data patterns, and (3) how to balance the influence of data and knowledge constraints while maintaining rule reliability during training. A critical difficulty is that not all logical rules are equally useful—some may be redundant or even harmful to the prediction task, leading to degraded performance if treated uniformly.

To address these challenges, we propose a two-stage neuro-symbolic learning strategy. First, a weighted axiom loss stabilizes the interplay between data-driven and logic-driven objectives during pretraining, preventing unreliable rules from dominating the optimization. Second, a rule pruning mechanism filters out non-contributive rules based on satisfaction dynamics, producing a refined knowledge base for fine-tuning. This approach leverages the Logic Tensor Network framework~\cite{bagase_2022} to create a unified neuro-symbolic classifier that maintains high predictive accuracy while ensuring adherence to domain constraints.

Our key contributions are:
\begin{enumerate}
\item To introduce a principled approach for extracting and categorizing process knowledge into control-flow, temporal, and payload features, formalizing them using Linear Temporal Logic (LTL) and first-order logic for integration into neural predictive models.

\item To propose a novel training strategy combining weighted axiom loss and rule pruning that addresses a fundamental limitation of Logic Tensor Networks (i.e., their tendency to satisfy logical formulas at the expense of predictive accuracy), resulting in more stable training and superior performance even with limited compliant examples.
\end{enumerate}

The paper is structured as follows. \autoref{sec:soa} reviews related work in predictive process monitoring and neuro-symbolic AI. \autoref{sec:preliminaries} introduces preliminary concepts. \autoref{sec:approach} details our knowledge formalization and rule pruning mechanism. \autoref{sec:evaluation} presents comprehensive experimental results, and \autoref{sec:conclusion} concludes with future directions.

\section{Related Work}
\label{sec:soa}
Predictive process monitoring (PPM) aims to forecast process outcomes and future events from event logs.
Deep learning approaches have significantly advanced this field, with LSTM networks~\cite{DBLP:conf/caise/TaxVRD17,DBLP:journals/dss/EvermannRF17}, Transformer architectures~\cite{DBLP:journals/corr/abs-2104-00721}, bidirectional models~\cite{DBLP:conf/bpm/0001DR19,DBLP:conf/icws/WangYLS19}, and topological approaches~\cite{DBLP:conf/kdd/HuangXH25} demonstrating strong performance in capturing complex sequential patterns.

However, most approaches remain purely data-driven, ignoring domain constraints crucial for compliance-critical settings.
Few studies have integrated explicit process knowledge: Vazifehdoostirani et al.~\cite{vazifehdoostirani2022encoding} encode control-flow patterns as static input features, while Di Francescomarino et al.~\cite{di2017eye} enforce LTL constraints as hard post-processing rules. Mezini et al.~\cite{DBLP:journals/corr/abs-2509-00834} incorporate LTL constraints in the training process using a differentiable loss function to address the problem of suffix prediction.

These approaches have three key limitations: (1) focusing primarily on control-flow constraints while neglecting temporal and payload knowledge, (2) injecting knowledge either at input or output but not throughout the learning process, and (3) lacking mechanisms to manage conflicting or noisy rules.

Neuro-symbolic AI addresses these limitations by integrating neural learning with symbolic reasoning~\cite{bhrato_2024}.  Several frameworks have emerged for this purpose. Semantic loss functions~\cite{DBLP:conf/icml/XuZFLB18} and constraint-aware training methods~\cite{DBLP:conf/icml/FischerBDGZV19} translate logical formulas into differentiable loss terms. 
Gradient-based approaches like RL-Net~\cite{DBLP:conf/pakdd/DierckxVN23} learn interpretable ordered rules by combining neural network optimization with symbolic rule representation. Systems like $\delta$ILP~\cite{evgr_2018} and DeepProbLog~\cite{maduki_2018} combine logic programming with deep learning. In the temporal domain, methods like TempRule~\cite{DBLP:conf/pakdd/BaoWWWZZFW25} demonstrate the effectiveness of learning symbolic rules on knowledge graphs. Knowledge-infusion techniques~\cite{DBLP:conf/pakdd/ZhaoJH20,DBLP:conf/pakdd/YingLYZYGL24,DBLP:conf/pakdd/LiWLLM25} fuse knowledge into neural architectures through attention mechanisms and learnable constraints.

Among these frameworks, Logic Tensor Networks (LTNs)~\cite{bagase_2022} are particularly well-suited for predictive process monitoring. Unlike approaches that require manual translation of logical constraints into loss terms or architectural modifications, LTNs provide a unified framework that directly embeds first-order logic into differentiable computation graphs using fuzzy semantics. This enables three key advantages for our domain: (1) natural representation of diverse process constraints, such as control-flow patterns expressed in LTL, temporal constraints on durations and waiting times, and payload-based business rules, within a single formalism; (2) soft constraint satisfaction through fuzzy logic semantics, which gracefully handles partial compliance and conflicting rules common in real-world process data; and (3) seamless integration of reasoning and learning, where logical formulas actively guide neural optimization rather than serving as post-hoc corrections.

Building on the LTN framework, our work makes two key contributions. First, we demonstrate how to systematically formalize complex business rules and temporal constraints from process mining using FOL and LTL for integration into neural predictive models. Second, we introduce a two-stage optimization strategy that addresses a fundamental limitation of LTNs (i.e., their tendency to prioritize logical consistency over predictive accuracy) through a weighted axiom loss during pretraining and a rule pruning mechanism that filters out inconsistent or non-contributive rules, resulting in more stable and interpretable training dynamics.

\section{Preliminaries}
\label{sec:preliminaries}
\subsection{Logic Tensor Networks}
\label{sec:ltn}

\begin{figure*}[t]
\centering
\includegraphics[width=0.95\textwidth]{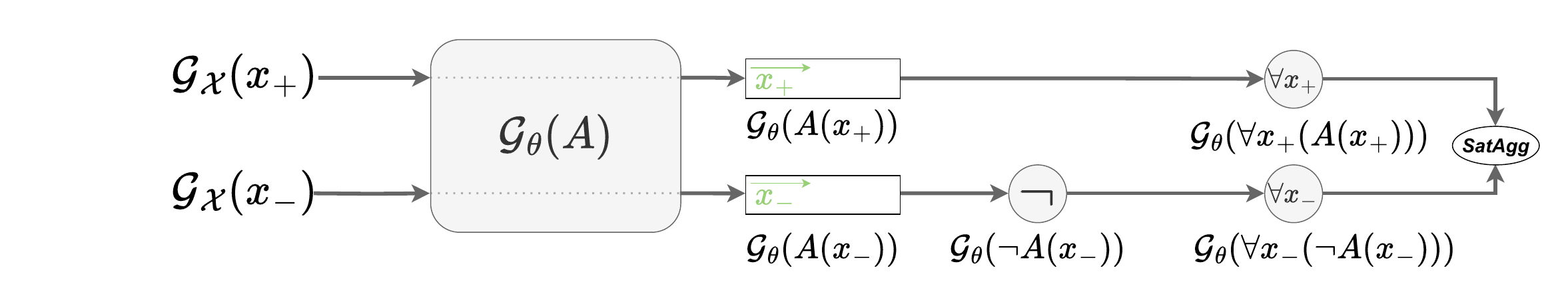}
\caption{Binary classification using LTN: loan applications (constants $\mathcal{C}$) are evaluated by predicate $A$ (neural network) to determine approval. Variables $x_+$ and $x_-$ represent positive and negative classes, aggregated by universal quantifiers $\forall$ to compute satisfaction.} \label{fig:fol-ltn}
\end{figure*}

Logic Tensor Networks (LTNs)~\cite{bagase_2022} implement Real Logic, extending First-Order Logic (FOL) with fuzzy semantics where truth values are continuous in $[0,1]$. This enables differentiable integration of logical constraints with neural learning.

\paragraph{Real Logic Components.} LTN comprises: (1) \textit{constants} $\mathcal{C}$ representing individuals (e.g., loan applications), (2) \textit{variables} $\mathcal{X}$ as typed placeholders (e.g., $x_+$ for approved applications), (3) \textit{functions} $\mathcal{F}$ computing derived features, (4) \textit{predicates} $\mathcal{P}$ expressing properties (e.g., $A$ for "approved"), and (5) logical connectives $\{\neg,\wedge,\vee,\rightarrow,\leftrightarrow\}$ and quantifiers $\{\forall, \exists\}$.

\paragraph{Grounding and Differentiability.} Each logic element is \textit{grounded} to real tensors via function $\mathcal{G}$. Crucially, predicates are implemented as neural networks outputting truth values in $[0,1]$, enabling gradient-based learning. Quantifiers are realized through differentiable aggregators:
\[\forall x: \mathrm{pMeanError}(u_1,...,u_n)=1-\left(\frac{1}{n}\sum_{i=1}^n (1-u_i)^p\right)^{\frac{1}{p}}, \
\exists x: \mathrm{pMean}(u_1,...,u_n)=\left(\frac{1}{n}\sum_{i=1}^n u_i^p\right)^{\frac{1}{p}}\]
where $p\geq1$ controls strictness. For binary classification (illustrated in \autoref{fig:fol-ltn}), the knowledge base contains: $\forall x_+ A(x_+)$ (all positive examples satisfy $A$) and $\forall x_- \neg A(x_-)$ (all negative examples falsify $A$). The neural predicate $A$ learns to output values near 1 for approved applications and near 0 for rejected ones.

\paragraph{Learning Objective.} LTN minimizes: $L = 1 - \underset{\phi \in \mathcal{K}}{\mathrm{SatAgg}}\mathcal{G_\theta}(\phi)$
where $\mathcal{K}$ is the knowledge base, $\mathrm{SatAgg}$ aggregates satisfaction degrees across axioms (using pMeanError), and $\theta$ denotes neural parameters. This objective ensures predictions satisfy both data labels and logical constraints through continuous optimization.

\subsection{Predictive Process Monitoring}
\label{sec:ppm}

An \textit{event} $e=(a, c, t, attr_1,...,attr_n)$ comprises activity $a$, case identifier $c$, timestamp $t$, and attributes. A \textit{trace} $\sigma=\langle e_1,...,e_n\rangle$ is an ordered sequence with identical case identifiers and non-decreasing timestamps. A \textit{trace prefix} $l=\langle e_1,...,e_k\rangle$ represents the first $k$ events of a trace.

Let $\mathcal{L}$ denote all prefixes, with $\mathcal{L}_+$ and $\mathcal{L}_-$ representing positive and negative classes. A \textit{predictive model} $\Delta: \mathcal{L}\rightarrow\{+,-\}$ classifies prefixes into outcome classes.

\paragraph{LTN Mapping.} Process monitoring naturally maps to Real Logic: variables $\mathcal{X}$ correspond to trace prefixes, constants $\mathcal{C}$ to activities and attribute values, functions $\mathcal{F}$ to temporal computations (e.g., time elapsed between activities) and aggregations (e.g., patient age, avg. requested amount), and predicates $\mathcal{P}$ to predictive models and constraints. For example, the rule "antibiotics not administered within two hours after surgery for elderly patients leads to complications" formalizes as:
\[\forall l\left((WaitTime(l,\texttt{Surg},\texttt{ATB})>2\wedge Age(l)>60)\rightarrow P(l)\right)\]
where $WaitTime, Age \in \mathcal{F}$ and $P \in \mathcal{P}$ models complication risk.

\section{Neuro-Symbolic AI for Predictive Process Monitoring}
\label{sec:approach}

\begin{figure}[t]
\centering
\includegraphics[width=0.8\textwidth]{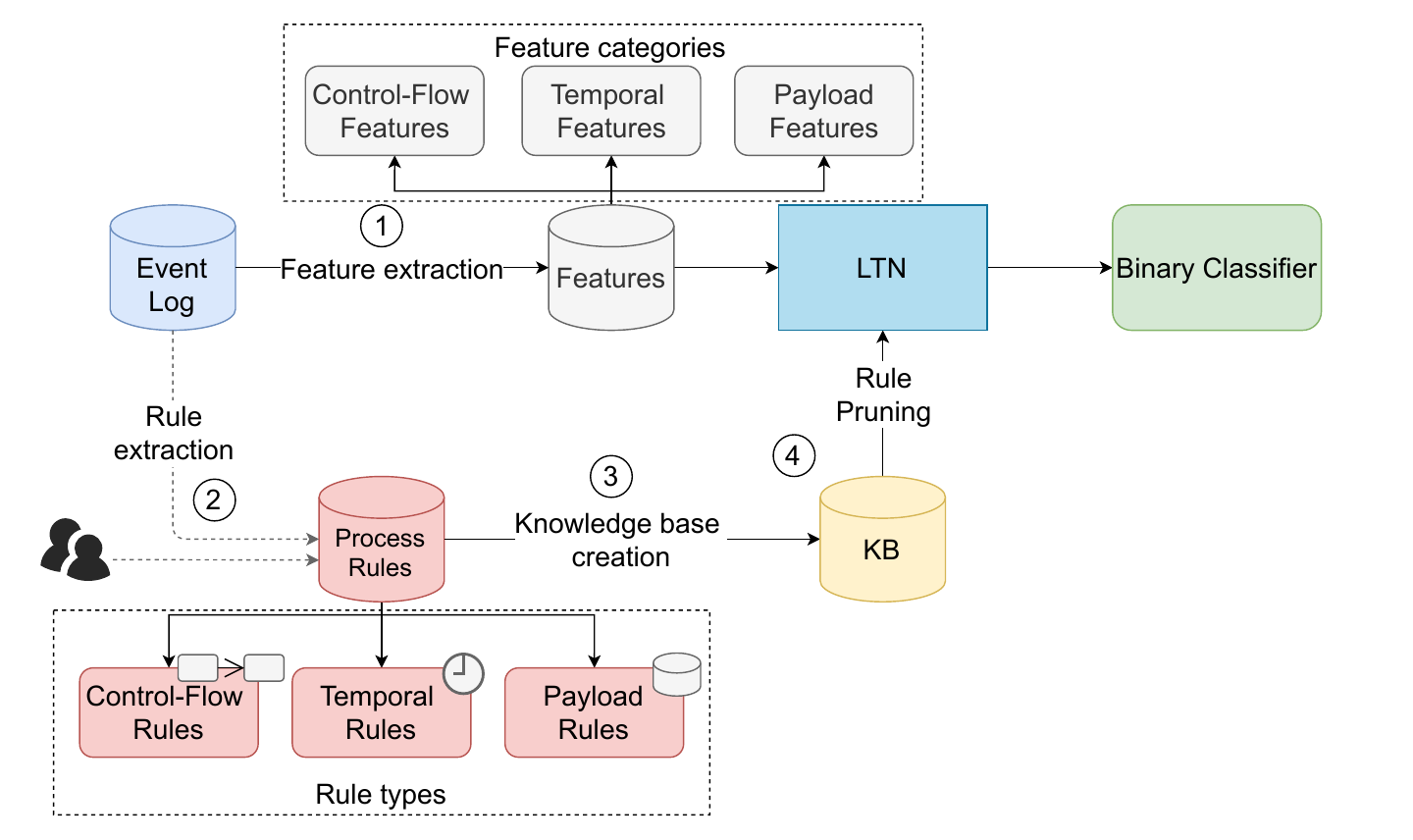}
\caption{The pipeline followed in the approach, which consists of feature extraction, rule extraction, knowledge base creation and injection, rule pruning, and then the creation of the neuro-symbolic model leveraging the LTN framework.} \label{fig:pipeline2}
\end{figure}

Our approach integrates domain knowledge into predictive process monitoring through the pipeline in \autoref{fig:pipeline2}. We extract features and rules from event logs, construct a knowledge base in FOL, and inject it into an LTN-based binary classifier.

Consider predicting post-surgical complications in a healthcare process with activities: medical history review ($\texttt{Rev}$), physical examination ($\texttt{Exam}$), laboratory tests ($\texttt{Lab}$), antibiotic administration ($\texttt{ATB}$), surgery ($\texttt{Surg}$), pain management ($\texttt{PAdm}$), and postoperative check-up ($\texttt{PostCU}$). We use predicate $P(l)$ to model complication probability for prefix $l$, with $l_+$ and $l_-$ denoting positive and negative classes.

\subsection{Feature Extraction}
\label{sec:featureextraction}
We extract three feature categories that define the vocabulary for rule formalization:
\begin{compactitem}
    \item \textit{Control-flow features}: Sequential structures (precedence, repetition, branching). Example: $IsNext(l, \texttt{Rev}, \texttt{Exam})$.
    \item \textit{Temporal features}: Time-based relationships (durations, waiting times, cycle times). Example: $WaitTime(l, \texttt{Surg},\texttt{ATB})$.
    \item \textit{Payload features}: Case-level (static attributes) and event-level (activity-specific attributes) contextual data.
\end{compactitem}

\subsection{Rule Extraction}
\label{sec:knowledgeextraction}

Rules are extracted automatically via process mining techniques or manually by domain experts. We extract three rule types aligned with the feature space:
\paragraph{Control-flow rules.} Sequential execution constraints represented using Linear Temporal Logic (LTL). We employ Declare mining~\cite{mamova_2011} to identify constraints. Examples:
\begin{compactitem}
    \item $\square(\texttt{Rev} \implies \lozenge \texttt{Exam})$: if $\texttt{Rev}$ occurs, $\texttt{Exam}$ must follow
    \item $\square(\texttt{Surg}\implies \bigcirc \texttt{PostCU})$: $\texttt{PostCU}$ immediately follows $\texttt{Surg}$
\end{compactitem}

\paragraph{Temporal rules.} Timing constraints extracted from SLA compliance analysis~\cite{rafava_2013}, expressed as IF-THEN statements. Example: "if $\texttt{ATB}$ occurs within two hours after $\texttt{Surg}$, complications decrease."

\paragraph{Payload rules.} Statistical correlations~\cite{caeppu_2018} between attributes and outcomes, expressed as IF-THEN statements. Example: "if oxygen saturation $<90\%$ post-surgery, complication risk increases."

\subsection{Knowledge Base Construction}

\paragraph{FOL Translation.} We convert extracted rules to FOL:
\begin{compactitem}
    \item \textit{Control-flow}: LTL-to-FOL conversion~\cite{deva_2013} preserves temporal semantics. Example: $\square(\texttt{Rev} \implies \lozenge \texttt{Exam})$ translates to $\forall l(HasAct(l, \texttt{Rev}) \wedge Next(l, \texttt{Rev},\texttt{Exam}))$.
    \item \textit{Temporal/Payload}: IF-THEN rules decompose into antecedent-consequent formulas. Example: $\forall l_+(WaitTime(l_+,\texttt{Surg},\texttt{ATB}) \leq 2 \rightarrow \neg P(l_+))$.
\end{compactitem}

\paragraph{Axiom Types.} In our framework, we distinguish between two types of axioms in the knowledge base: data axioms $\mathcal{K_D}$ and knowledge axioms $\mathcal{K_P}$. The knowledge base comprises:
\begin{compactitem}
  \item \textit{Data axioms} $\mathcal{K_D}$: $\{\forall x_+ P(x_+), \forall x_- \neg P(x_-)\}$: Classification-related constraints
  \item \textit{Knowledge axioms} $\mathcal{K_P}$: Domain-derived logical rules
\end{compactitem}

\subsection{Two-Stage Optimization with Rule Pruning}

\begin{algorithm}[!t]
\caption{Two-stage optimization with rule pruning for compliance-aware scenarios}\label{algpruning}
\begin{algorithmic}[1]
\REQUIRE $\alpha+\beta=1$, thresholds $\tau$, $\lambda$
\STATE Initialize $\theta$, $\mathcal{K_D}$, $\mathcal{K_P}=\{r_1,\dots,r_n\}$
\COMMENT{Phase 1: Weighted pretraining and pruning}
\FOR{epoch $=1$ to $E_p$}
    \STATE $L \gets 1- (\alpha\cdot\mathrm{SatAgg}(\mathcal{K_D}) + \beta\cdot\mathrm{SatAgg}(\mathcal{K_P}))$
    \STATE Update $\theta$ via backpropagation
\ENDFOR
\STATE Initialize $\mathcal{K'_P} \gets \emptyset$
\FOR{each $r_i \in \mathcal{K_P}$}
    \STATE Compute $g_i \gets \bar{s_i}\cdot e^{-\lambda\cdot \mathrm{Var}(s_i)}$
    \IF{$g_i \geq \tau$}
        \STATE Add $r_i$ to $\mathcal{K'_P}$
    \ENDIF
\ENDFOR
\COMMENT{Phase 2: Fine-tuning on pruned knowledge base}
\FOR{epoch $=1$ to $E_f$}
    \STATE $L \gets 1 - \mathrm{SatAgg}(\mathcal{K_D}\cup\mathcal{K'_P})$
    \STATE Update $\theta$ via backpropagation
\ENDFOR
\end{algorithmic}
\end{algorithm}

To address LTNs' tendency to satisfy logical constraints at the expense of predictive accuracy, we propose a two-stage optimization strategy (\autoref{algpruning}).

\paragraph{Phase 1: Weighted Pretraining and Rule Selection.} We train using weighted loss:
\begin{equation}
L = 1 - \left(\alpha\cdot\underset{\phi \in \mathcal{K_D}}{\mathrm{SatAgg}}\mathcal{G_\theta}(\phi)
+ \beta\cdot\underset{\phi \in \mathcal{K_P}}{\mathrm{SatAgg}}\mathcal{G_\theta}(\phi)\right)
\label{eq:weighted_loss}
\end{equation} Setting $\alpha>\beta$ prioritizes data learning, preventing vacuous satisfaction of knowledge axioms before the classification predicate is properly learned. After $E_p$ epochs, we construct a refined knowledge base $\mathcal{K'_P}$ by selecting rules via gating scores obtained on the validation set:
\[g_i=\bar{s_i}\cdot e^{-\lambda\cdot \mathrm{Var}(s_i)}\]
where $\bar{s_i}$ and $\mathrm{Var}(s_i)$ denote mean and variance of rule $r_i$'s satisfaction across samples. Rules with $g_i \geq \tau$ are added to $\mathcal{K'_P}$, retaining only consistent, contributive axioms.

\paragraph{Phase 2: Fine-tuning.} The model is fine-tuned on the refined knowledge base $\mathcal{K_D}\cup\mathcal{K'_P}$ for $E_f$ epochs using \autoref{eq:weighted_loss} (with $\mathcal{K'_P}$ instead of $\mathcal{K_P}$) in the compliance-aware scenario, while for the temporal plit scenario we use the standard LTN loss $L = 1 - \mathrm{SatAgg}(\mathcal{K_D}\cup\mathcal{K'_P})$.

\paragraph{Hyperparameters.} We set $\alpha=0.8$, $\beta=0.2$, $E_p=5$, $E_f=50$, $\lambda\in\{1,2\}$ based on dataset characteristics, and select $\tau$ via validation. Gating scores are computed analytically, avoiding learnable parameters and preserving interpretability.

\section{Evaluation}
\label{sec:evaluation}

We evaluate our approach\footnote{The code to reproduce our experiments is available at \url{https://github.com/FabrizioDeSantis/NeSy-PPM}.} by addressing three research questions:
\begin{itemize}
\item \textbf{RQ1:} Does injecting domain knowledge as logical constraints improve predictive performance compared to purely data-driven approaches?
\item \textbf{RQ2:} What is the contribution of the two-stage optimization with rule pruning mechanism to model performance?
\item \textbf{RQ3:} How effectively does the approach generalize to compliance-constrained scenarios with limited compliant training examples?
\end{itemize}

\subsection{Experimental Setup}
\label{sec:experimentalsetup}

\paragraph{Model Variants.} To address RQ1 and RQ2, we compare:
\begin{compactitem}
\item \textbf{LSTM~\cite{DBLP:conf/caise/TaxVRD17}, TFR~\cite{DBLP:journals/corr/abs-2104-00721}}: Purely data-driven baselines with binary cross-entropy loss (no logical components)
\item \textbf{LTN-Data-L, LTN-Data-T}: LTN~\cite{bagase_2022} with LSTM/Transformer backbones trained only on data axioms $\mathcal{K_D}$ (no domain knowledge)
\item \textbf{LTN-NoP-L, LTN-NoP-T}: LTN with data and knowledge axioms $\mathcal{K_D}\cup\mathcal{K_P}$ but without weighted loss or pruning
    \item \textbf{Two-Stage-L, Two-Stage-T}: Our complete approach with weighted axiom loss and rule pruning
\end{compactitem}

\paragraph{Architecture and Training.} LSTM: 2 layers, hidden size 128. Transformer: 2 attention heads, hidden size 128. Both use Adam optimizer (lr=0.001, batch size=32). Categorical features use embeddings with a dimension of 32; numerical features are standardized and passed through linear layers. We use simplified architectures to isolate knowledge integration effects, though our approach is architecture-agnostic. All results report mean and standard deviation over 5 random seeds.

\paragraph{Evaluation Protocols.} We employ two evaluation settings:
\begin{compactitem}
\item \textit{Temporal split}: Standard 80-20 train-test split based on completion timestamps, evaluating predictive performance on future cases (addresses RQ1 and RQ2)
\item \textit{Compliance-aware split}: Test set enriched with rule-compliant traces (satisfying logical rules and labels) plus random samples, assessing generalization to compliance-constrained scenarios (addresses RQ3)
\end{compactitem}

\subsection{RQ1: Impact of Domain Knowledge Injection}

To answer RQ1, we compare purely data-driven baselines (LSTM~\cite{DBLP:conf/caise/TaxVRD17}, TFR~\cite{DBLP:journals/corr/abs-2104-00721}) against knowledge-enhanced variants on the temporal split (\autoref{tab:rq1res}). 

\begin{table}[!t]
\centering
\caption{RQ1: Performance on temporal split comparing purely data-driven baselines with knowledge-enhanced variants. Results show mean and standard deviation over 5 seeds.}
\label{tab:rq1res}
\resizebox{\columnwidth}{!}{%
\begin{tabular}{ccccccccc}
\hline
\multirow{2}{*}{\textbf{Method}} &
  \multicolumn{2}{c}{\textbf{Sepsis}} &
  \multicolumn{2}{c}{\textbf{BPIC2012}} &
  \multicolumn{2}{c}{\textbf{BPIC2017}} &
  \multicolumn{2}{c}{\textbf{Traffic fines}} \\ \cline{2-9} 
              & Accuracy     & F1           & Accuracy     & F1           & Accuracy     & F1           & Accuracy     & F1           \\ \hline
\textbf{LSTM} & 88.69$\pm$0.92 & 70.59$\pm$1.31 & 52.47$\pm$0.77 & 52.44$\pm$0.74 & 70.16$\pm$0.41 & 69.47$\pm$0.44 & 67.34$\pm$0.42 & 66.93$\pm$0.28 \\
\textbf{TRF}  & 90.11$\pm$0.47 & 73.80$\pm$0.36 & 52.29$\pm$0.58 & 51.86$\pm$0.92 & 71.15$\pm$0.25 & 70.74$\pm$0.22 & 67.81$\pm$0.11 & 67.34$\pm$0.14 \\
\textbf{Two-Stage-L} &
  89.39$\pm$0.47 &
  \textbf{75.82$\pm$0.92} &
  60.33$\pm$0.86 &
  57.25$\pm$0.38 &
  70.69$\pm$0.89 &
  70.67$\pm$0.89 &
  \textbf{68.44$\pm$0.43} &
  \textbf{68.20$\pm$0.38} \\
\textbf{Two-Stage-T} &
  \textbf{91.30$\pm$0.11} &
  75.63$\pm$0.99 &
  \textbf{65.27$\pm$0.27} &
  \textbf{62.12$\pm$0.39} &
  \textbf{72.30$\pm$0.11} &
  \textbf{72.16$\pm$0.16} &
  67.94$\pm$0.11 &
  67.72$\pm$0.08 \\ \hline
\end{tabular}%
}
\end{table}

\textit{Key Findings.} Two-Stage-L and Two-Stage-T consistently outperform their purely neural counterparts across all datasets. Improvements are most pronounced on smaller datasets: \textit{Sepsis} shows F1 improvements of 5.23\% (LSTM$\to$Two-Stage-L) and 1.83\% (TFR$\to$Two-Stage-T), while \textit{BPIC2012} exhibits gains of 4.81\% and 10.26\% respectively. For larger datasets like \textit{Traffic fines}, improvements are more modest (1.1\% and 0.38\%), as abundant training data reduces the relative contribution of logical rules.

Architecture comparison reveals no clear overall winner: LSTMs excel on short sequences (\textit{Sepsis}, \textit{Traffic fines}: max length 13 and 10), achieving 75.82\% and 68.20\% F1 with Two-Stage-L, while Transformers perform better on longer traces (\textit{BPIC2012}, \textit{BPIC2017}: max length 40 and 20), reaching 62.12\% and 72.16\% F1 with Two-Stage-T.

\subsection{RQ2: Ablation Study on Two-Stage Optimization}

To answer RQ2, we conduct an ablation study comparing: (1) LTN-Data (data axioms only), (2) LTN-NoP (knowledge axioms without pruning), and (3) Two-Stage (our complete method) on the temporal split (\autoref{tab:rq2res}).

\begin{table}[!t]
\centering
\caption{RQ2: Ablation study comparing LTN variants on temporal split: Data-only (no knowledge), NoP (knowledge without pruning), and Two-Stage (complete method).}
\label{tab:rq2res}
\resizebox{\columnwidth}{!}{%
\begin{tabular}{ccccccccc}
\hline
\multirow{2}{*}{\textbf{Method}} &
  \multicolumn{2}{c}{\textbf{Sepsis}} &
  \multicolumn{2}{c}{\textbf{BPIC2012}} &
  \multicolumn{2}{c}{\textbf{BPIC2017}} &
  \multicolumn{2}{c}{\textbf{Traffic fines}} \\ \cline{2-9} 
 &
  Accuracy &
  F1 &
  Accuracy &
  F1 &
  Accuracy &
  F1 &
  Accuracy &
  F1 \\ \hline
\textbf{LTN-Data-L} &
  82.56$\pm$4.52 &
  69.10$\pm$4.22 &
  54.24$\pm$0.63 &
  53.18$\pm$0.63 &
  69.73$\pm$0.07 &
  69.71$\pm$0.07 &
  67.59$\pm$0.60 &
  67.31$\pm$0.56 \\
\textbf{LTN-Data-T} &
  81.09$\pm$6.25 &
  66.79$\pm$4.65 &
  57.17$\pm$1.13 &
  55.40$\pm$0.77 &
  71.15$\pm$0.25 &
  70.74$\pm$0.22 &
  67.53$\pm$0.19 &
  67.27$\pm$0.1 \\
\textbf{LTN-NoP-L} &
  21.36$\pm$5.96 &
  18.79$\pm$2.95 &
  54.50$\pm$4.35 &
  45.00$\pm$5.54 &
  69.16$\pm$0.24 &
  66.27$\pm$0.36 &
  59.33$\pm$0.27 &
  46.35$\pm$0.41 \\
\textbf{LTN-NoP-T} &
  21.69$\pm$7.58 &
  19.18$\pm$4.76 &
  47.47$\pm$6.68 &
  33.82$\pm$2.70 &
  68.54$\pm$0.52 &
  66.09$\pm$0.69 &
  57.57$\pm$0.86 &
  41.78$\pm$1.94 \\
\textbf{Two-Stage-L} &
  89.39$\pm$0.47 &
  \textbf{75.82$\pm$0.92} &
  60.33$\pm$0.86 &
  57.25 $\pm$0.38 &
  70.69$\pm$0.89 &
  70.67$\pm$0.89 &
  \textbf{68.44$\pm$0.43} &
  \textbf{68.20$\pm$0.38} \\
\textbf{Two-Stage-T} &
  \textbf{91.30$\pm$0.11} &
  75.63$\pm$0.99 &
  \textbf{65.27$\pm$0.27} &
  \textbf{62.12$\pm$0.39} &
  \textbf{72.30$\pm$0.11} &
  \textbf{72.16$\pm$0.16} &
  67.94$\pm$0.11 &
  67.72$\pm$0.08 \\ \hline
\end{tabular}%
}
\end{table}

\textit{Key Findings.} The results reveal a critical insight: adding knowledge axioms without proper control (LTN-NoP) severely degrades performance, particularly on small datasets. On \textit{Sepsis}, LTN-NoP-L and LTN-NoP-T achieve only 18.79\% and 19.18\% F1, substantially worse than even purely data-driven baselines. This occurs because LTNs optimize overall knowledge base satisfaction, allowing models to exploit logical shortcuts (vacuous satisfaction) rather than learning meaningful predictive patterns.

Our two-stage approach addresses this fundamental limitation. On \textit{Sepsis}, Two-Stage-L recovers to 70.59\% F1 (51.80\% improvement over LTN-NoP-L). Similar patterns emerge across all datasets: \textit{BPIC2012} shows 28.12\% improvement, \textit{BPIC2017} shows 5.78\% improvement, and \textit{Traffic fines} shows 25.94\% improvement.

Comparing LTN-Data to Two-Stage isolates the net benefit of domain knowledge when properly integrated. Two-Stage consistently outperforms LTN-Data: improvements range from 10.26\% on \textit{BPIC2012} to 8.84\% on \textit{Sepsis}, demonstrating that the two-stage mechanism successfully harnesses knowledge benefits while avoiding pitfalls.

\subsection{RQ3: Generalization to Compliance Constraints}

To answer RQ3, we evaluate on the compliance-aware test set that combines rule-compliant traces with random samples (\autoref{tab:rq3res}). This setting is particularly challenging because: (1) compliant examples in training are scarce (4\% in \textit{Sepsis}, 18.25\% in \textit{BPIC2012}), and (2) the test set is enriched with compliant cases (32.3\% and 61.9\% respectively), requiring models to generalize beyond the training distribution.

\begin{table}[!t]
\centering
\caption{RQ3: Performance on compliance-aware test set. Models must generalize to scenarios emphasizing rule-compliant behavior with limited compliant training examples.}
\label{tab:rq3res}
\resizebox{\columnwidth}{!}{%
\begin{tabular}{ccccccccc}
\hline
\multirow{2}{*}{\textbf{Method}} &
  \multicolumn{2}{c}{\textbf{Sepsis}} &
  \multicolumn{2}{c}{\textbf{BPIC2012}} &
  \multicolumn{2}{c}{\textbf{BPIC2017}} &
  \multicolumn{2}{c}{\textbf{Traffic fines}} \\ \cline{2-9} 
 &
  Accuracy &
  F1 &
  Accuracy &
  F1 &
  Accuracy &
  F1 &
  Accuracy &
  F1 \\ \hline
\textbf{LSTM} &
  81.50$\pm$2.41 &
  73.26$\pm$3.28 &
  53.89$\pm$1.59 &
  53.71$\pm$1.49 &
  65.71$\pm$1.04 &
  65.41$\pm$1.19 &
  77.47$\pm$0.21 &
  76.92$\pm$0.33 \\
\textbf{TRF} &
  82.74$\pm$1.39 &
  74.66$\pm$1.77 &
  56.16$\pm$0.79 &
  55.61$\pm$0.68 &
  68.77$\pm$0.83 &
  68.54$\pm$0.96 &
  77.39$\pm$0.48 &
  76.85$\pm$0.45 \\
\textbf{LTN-NoP-L} &
  39.91$\pm$4.83 &
  37.52$\pm$6.51 &
  47.58$\pm$5.82 &
  34.61$\pm$4.53 &
  65.28$\pm$0.05 &
  64.48$\pm$0.02 &
  64.93$\pm$0.05 &
  42.39$\pm$0.11 \\
\textbf{LTN-NoP-T} &
  34.55$\pm$4.10 &
  30.38$\pm$5.47 &
  63.61$\pm$0.74 &
  53.39$\pm$1.00 &
  62.02$\pm$6.69 &
  60.68$\pm$7.77 &
  65.15$\pm$0.09 &
  42.74$\pm$0.22 \\
\textbf{Two-Stage-L} &
  \textbf{91.47$\pm$2.82} &
  \textbf{89.96$\pm$2.96} &
  62.88$\pm$0.80 &
  60.07$\pm$0.49 &
  75.53$\pm$0.35 &
  74.66$\pm$0.22 &
  78.36$\pm$0.52 &
  77.76$\pm$0.59 \\
\textbf{Two-Stage-T} &
  89.21$\pm$0.88 &
  87.36$\pm$0.89 &
  \textbf{68.17$\pm$0.14} &
  \textbf{64.23$\pm$0.06} &
  \textbf{77.16$\pm$0.24} &
  \textbf{75.47$\pm$0.03} &
  \textbf{78.86$\pm$0.26} &
  \textbf{78.44$\pm$0.13} \\ \hline
\end{tabular}%
}
\end{table}

\textit{Key Findings.} Two-Stage variants substantially outperform baselines on this evaluation. On \textit{Sepsis} with only 4\% compliant training examples, Two-Stage-L achieves 89.96\% F1 compared to 73.26\% for LSTM (16.7\% improvement). The gap widens further compared to LTN-NoP-L (52.44\%), demonstrating that naive knowledge injection fails catastrophically in compliance-constrained settings.

\textit{Sepsis} (4\% compliant) shows the largest improvement (16.7\%), followed by \textit{BPIC2012} (18.25\%, 10.52\%), \textit{BPIC2017} (62.7\%, 10.06\%), and \textit{Traffic fines} (39.67\%, 1.52\%). This trend confirms that logical rules provide crucial inductive bias when compliant examples are scarce. Notably, all purely data-driven approaches (LSTM, TFR, LTN-Data) exhibit similar performance, unable to extrapolate compliance patterns from limited examples.

\subsection{Discussion}

Our evaluation provides several insights. For RQ1, domain knowledge injection consistently improves performance, with gains varying across datasets. The improvements are most pronounced under limited training data, indicating that knowledge can constrain model learning when statistical evidence is insufficient. In data-rich settings, where patterns are already well captured, the marginal benefit naturally diminishes. The results for RQ2 emphasize that knowledge integration is not inherently beneficial. Simple incorporation can compromise the optimization process and significantly degrade performance. The proposed two-stage optimization is therefore essential, as it stabilizes training by balancing statistical objectives and logical constraints, highlighting the non-trivial interaction between symbolic knowledge and data-driven learning. Finally, for RQ3, the approach is particularly effective in compliance-constrained scenarios with scarce compliant examples. While purely data-driven models tend to overlook rare but semantically critical behaviors, explicit encoding preserves relevant patterns, addressing a key limitation of conventional predictive process monitoring.

\section{Conclusion}
\label{sec:conclusion}

We presented a neuro-symbolic approach for predictive process monitoring that systematically integrates domain knowledge as differentiable logical constraints using Logic Tensor Networks. Our key contribution is a two-stage optimization strategy that addresses a fundamental limitation of LTNs: their tendency to satisfy logical formulas at the expense of predictive accuracy. The approach combines weighted axiom loss during pretraining to prioritize data learning, followed by rule pruning that retains only consistent, contributive axioms based on satisfaction dynamics.

Our evaluation on four real-world event logs answered three research questions. First, domain knowledge injection improves predictive performance, especially with limited training data (RQ1). Second, the two-stage optimization is essential—without it, knowledge injection can severely degrade performance through vacuous satisfaction of logical constraints (RQ2). Third, the approach effectively generalizes to compliance-constrained scenarios with scarce compliant training examples (RQ3), addressing a critical gap in current predictive process monitoring methods.

In future work, we plan to formalize an explicit representation defining specific templates to automatically connect features with rules. Moreover, we aim to evaluate our neuro-symbolic approach on a case study where domain knowledge plays a critical role, such as in healthcare. Finally, we plan to extend the approach to more process-aware tasks, such as predictions related to next events and time-related aspects.

\section*{Acknowledgements}

F. De Santis is supported by the Italian Ministry of University and Research (MUR) under the National Recovery and Resilience Plan (NRRP), Mission 4, Component 1, Investment 4.1, CUP D91I23000080006, funded by the European Union - NextGenerationEU.

\bibliographystyle{splncs04}
\bibliography{mybib}

\end{document}